\documentclass[]{InspireOmni_template}

\usepackage[toc,page,header]{appendix}

\usepackage{minitoc}
\usepackage{cleveref} 
\usepackage{subcaption}
\usepackage{booktabs}
\usepackage{wrapfig}
\usepackage{amsmath}
\usepackage{amssymb}

\newcolumntype{L}[1]{>{\raggedright\let\newline\\\arraybackslash\hspace{0pt}}m{#1}}
\newcolumntype{R}[1]{>{\raggedleft\let\newline\\\arraybackslash\hspace{0pt}}m{#1}}

\newcommand{\ignore}[1]{}

\makeatletter
\DeclareRobustCommand\onedot{\futurelet\@let@token\@onedot}
\def\@onedot{\ifx\@let@token.\else.\null\fi\xspace}

\makeatother

\definecolor{MyBlue}{rgb}{0.46, 0.50, 0.61}
\definecolor{MyDarkBlue}{rgb}{0,0.08,0.8}
\definecolor{MyDarkGreen}{RGB}{45,155,45}
\definecolor{MyDarkRed}{rgb}{0.8,0.02,0.02}
\definecolor{MyOrange}{rgb}{1.0, 0.4, 0.2}
\definecolor{MyPurple}{RGB}{111,0,255}
\definecolor{MyRed}{rgb}{0.8,0.0,0.0}
\definecolor{MyGold}{rgb}{0.75,0.6,0.12}
\definecolor{MyDarkgray}{rgb}{0.66, 0.66, 0.66}
\definecolor{MyBrown}{rgb}{0.65, 0.16, 0.16}
\definecolor{MyMutedRose}{rgb}{0.58, 0.29, 0.35}
\definecolor{JiayuanColor}{rgb}{0.60,0.43,0.48}
\definecolor{erranColor}{rgb}{24, 40, 113}

\definecolor{citecolor}{HTML}{696FAD}

\definecolor{bggray}{HTML}{F5F5F5}
\definecolor{pvdblue}{HTML}{DAE8FC}
\definecolor{RoseQuartzBg}{HTML}{F7CAC9}
\definecolor{RoseQuartz}{HTML}{F5A798}
\definecolor{Serenity}{HTML}{92A8D1}
\definecolor{OrangeRed}{rgb}{1.0, 0.27, 0.0}
\definecolor{RoyalBlue}{cmyk}{1, 0.50, 0, 0}
\definecolor{Turquoise}{HTML}{0F4C81}
\definecolor{mint}{rgb}{0.24, 0.71, 0.54}
\definecolor{green}{rgb}{0.0, 0.120, 0.0}

\newdimen\abovecrulesep
\newdimen\belowcrulesep
\makeatletter
\patchcmd{\@@@cmidrule}{\aboverulesep}{\abovecrulesep}{}{}
\patchcmd{\@xcmidrule}{\belowrulesep}{\belowcrulesep}{}{}
\makeatother

\definecolor{mybluetitle}{HTML}{4B527E} %

\definecolor{codegreen}{HTML}{478058}%
\definecolor{codegray}{rgb}{0.5,0.5,0.5}
\definecolor{codepurple}{HTML}{4F5E80} %
\definecolor{backcolour}{rgb}{0.95,0.95,0.92}
\lstdefinestyle{mystyle}{
    backgroundcolor=\color{backcolour},
    commentstyle=\color{codegreen},
    keywordstyle=\color{magenta},
    numberstyle=\tiny\color{codegray},
    stringstyle=\color{codepurple},
    basicstyle=\ttfamily\scriptsize,
    breakatwhitespace=false,
    breaklines=true,
    captionpos=b,
    keepspaces=true,
    frame=none,
    numbersep=5pt,
    showspaces=false,
    showstringspaces=false,
    showtabs=false,
    tabsize=2
}

\newtcolorbox{promptbox}[2][]{
    enhanced, 
    breakable,
    center title,
    left*=0pt, right*=0pt,
    boxsep=2pt, left=5pt, right=5pt,
    skin first=enhanced,
    skin middle=enhanced,
    skin last=enhanced,
    colback  = backcolour,
    fonttitle=\bfseries\rmfamily,
    fontupper=\scriptsize,
    title={\footnotesize\strut{#2}},
    #1
    }

\newtcolorbox{onebox}[2][]{
    enhanced, 
    center title,
    left*=0pt, right*=0pt,
    boxsep=2pt, left=5pt, right=5pt,
    skin first=enhanced,
    skin middle=enhanced,
    skin last=enhanced,
    colframe = mybluetitle!90,
  colback  = mybluetitle!10,
    fonttitle=\bfseries\rmfamily\fontfamily{phv}\selectfont,
    title={\footnotesize\strut{#2}  \refstepcounter{subsubsection} \addcontentsline{toc}{subsubsection}{\string\numberline{\thesubsubsection}#2}
    },
    #1
    }

\title{HSC-VLA: Hierarchical Scene-Clearing for Robust Bimanual Manipulation in Dense Clutter}

\author[1,2]{Zhen Liu}
\author[1,2]{Xinyu Ning}
\author[2]{Zhe Hu}
\author[1]{XinXin Xie}
\author[1]{Yitong Liu}

\author[2,3, \dagger]{Zhongzhu Pu}

\affiliation[1]{Beijing University of Posts and Telecommunications}
\affiliation[2]{InspireOmni AI}
\affiliation[3]{Tsinghua University}

\abstract{
Modern Vision--Language--Action models often suffer from critical instruction-following failures in high-density manipulation environments, where task-irrelevant visual clutter dilutes attention, corrupts grounding, and substantially degrades performance in complex long-horizon scenarios. To overcome the representation bottleneck of monolithic end-to-end architectures, we propose HSC-VLA, a hierarchical framework that decouples high-level visual-semantic reasoning from low-level, high-frequency sensorimotor execution through an explicit scene-clearing abstraction.
HSC-VLA employs a high-level Brain to decompose long-horizon tasks and to generate task-specific scene masks that preserve task-relevant geometry while suppressing distractors. The filtered observations are then passed to a low-level Cerebellum, a diffusion-based policy that performs bimanual manipulation using only mask-filtered vision and proprioception.
Extensive experiments in densely cluttered supermarket shelves demonstrate that HSC-VLA achieves 86.7\% aggregate success under high-density clutter, surpassing the best monolithic baseline ($\pi_0$-Full FT at 34.3\%) by 52.4\%. HSC-VLA also exhibits strong long-horizon performance, reaching 72\% on clutter sorting and 66\% on restocking, demonstrating strong robustness and effective failure recovery in complex cluttered manipulation.
}

\date{March 8, 2026}

\correspondence{ \email{jaycening@inspireomni.ai}}

\begin{document}

\maketitle


\section{Introduction}

Automating large-scale manipulation in unstructured, high-density environments remains a fundamental challenge for autonomous systems deployed in logistics and service sectors. 
Such environments are characterized by massive, heterogeneous inventories and complex spatial layouts. Supermarkets, for instance, constitute a particularly rigorous scenario, requiring robots to manage thousands of distinct stock-keeping units (SKUs) within cluttered and dynamically evolving shelves ~\citep{soshin2025robobenchmart}. The resulting visual complexity, marked by severe occlusions, irregular object arrangements, and optical artifacts such as specular reflections from diverse packaging materials, substantially complicates perception of the situation~~\citep{kossira2025towards}.
Deploying bimanual robotic systems in the scenarios lead to the core challenges in extracting precise geometric affordances while reliably distinguishing task-critical objects from overwhelming environmental clutter ~\citep{huang2025prism,jiang2025rethinking,zhou2025you}. Failure to isolate relevant spatial structure leads to unstable grasping, misaligned placements, and compounding errors in long-horizon execution.

Recent advancements in Vision-Language-Action (VLA) models, including RT-2 ~\citep{rt22023arxiv}, pi0 ~\citep{black2024pi0}, and RDT-1B ~\citep{liu2024rdt}, have shown that multimodal transformers can map instructions and visual observations directly to motor commands in an end-to-end manner. 
However, these monolithic models often encounter a \textit{representation bottleneck} in high-density, cluttered environments~~\citep{ai2025review, kim2025freeson}. When high-dimensional raw pixels are encoded directly  into latent representations for action generation, task-relevant signals become entangled with complex and irrelevant background information, substantially increasing the perceptual burden of visual-semantic reasoning. This entanglement gives rise to an \textit{attention dilution effect}~~\citep{zhang2024selective, huang2025prism}, where model capacity is disproportionately allocated to modeling distractors rather than capturing the geometric structure required for precise manipulation. Consequently, even minor distribution shifts in clutter configuration can trigger catastrophic failures, as the policy lacks an explicit mechanism for isolating task-relevant geometry~~\citep{obeyed2025, wen2025diffusionvla}. Without structured perceptual abstraction, the model must implicitly perform scene disentanglement within its action space, an objective that scales poorly with environmental density and scene complexity.


\begin{figure}[!t]
    \centering
    \includegraphics[width=0.88\textwidth]{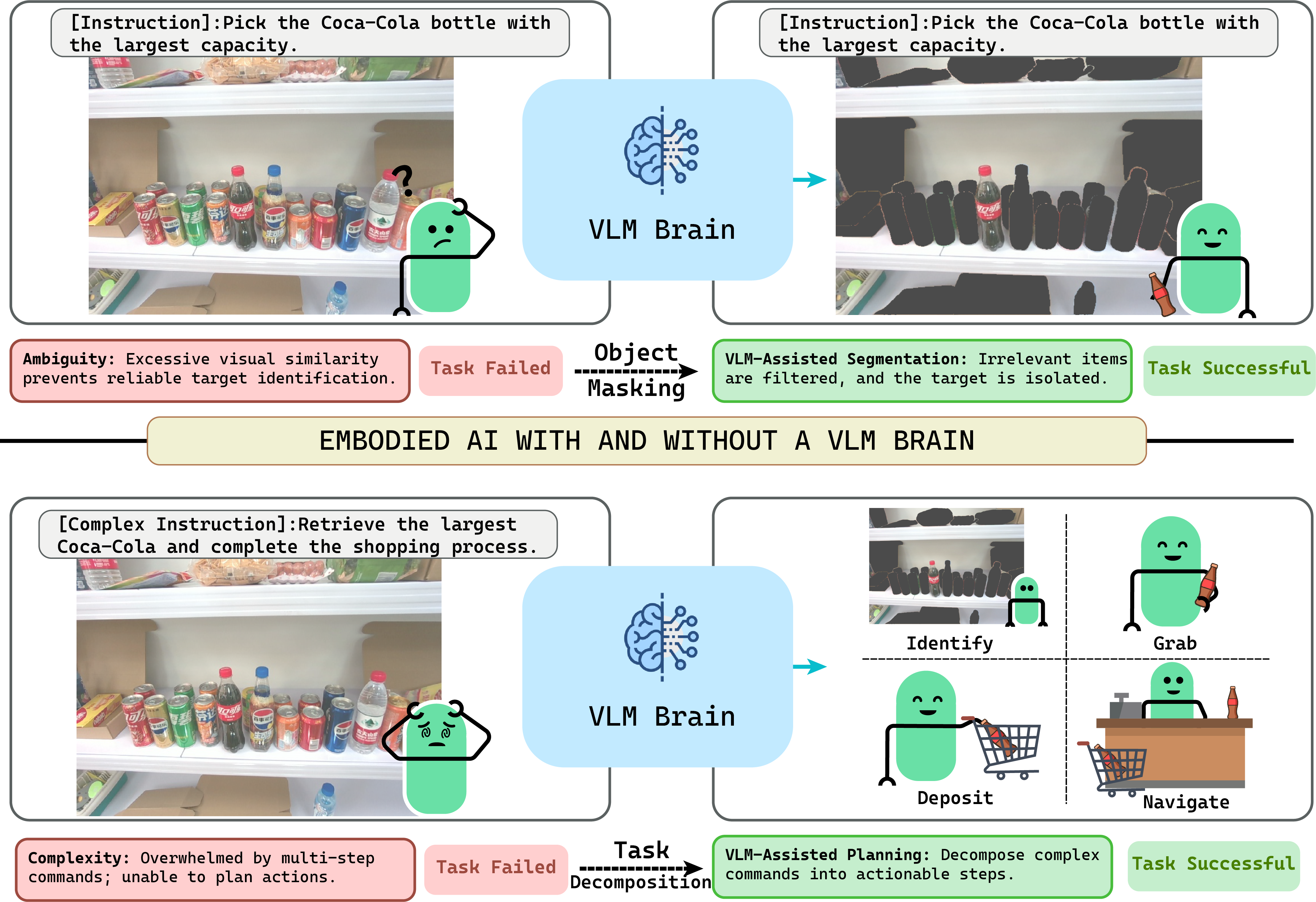} 
    
    \caption{How VLM Improves Perception and Planning in Embodied AI. 
    [Upper Row]: VLM-assisted segmentation reduces visual ambiguity by masking irrelevant objects, enabling precise target identification in cluttered environments. 
    [Lower Row]: VLM-assisted planning decomposes high-level, long-horizon instructions into clear, executable steps (Identify, Grasp, Place, and Navigate). 
    The VLM Brain functions as both a semantic filter for perception and a structured reasoner for decision-making.}
    
    \label{fig:scene}
\end{figure}


Beyond perceptual overload, current VLA models often struggle with complex tasks that involve multiple interdependent sub-tasks, a limitation rooted in the inherent tension between high-level reasoning and low-level control~~\citep{bubeck2023paper}. In unstructured environments like supermarkets, a seemingly simple instruction (e.g., ``restock the milk behind the juice boxes'') requires multi-stage causal reasoning, object relocation, and coordinated bimanual manipulation, as illustrated in Figure~\ref{fig:scene}. Monolithic VLAs typically learns planning, memory, and control within a single latent representation, often leading to causal confusion: sub-goals are incorrectly sequenced, recovery behaviors are poorly grounded, and policies become trapped in non-productive interaction loops when encountering physical resistance. Moreover, these systems struggle to maintain a persistent belief state, frequently ``forgetting'' occluded objects or prior progress during long-horizon execution. This gap between abstract intent and physical realization highlights the need for structural decoupling that separates semantic orchestration from high-frequency sensorimotor stabilization~~\citep{erdogan2025plan}.

To address these limitations, we introduce \textbf{HCS-VLA}, a hierarchical framework inspired by the functional separation between strategic reasoning and reactive execution ~\citep{wu2025hibernac, vijayaraghavan2025development}. Our approach decouples long-horizon planning from low-level control via an explicit \textbf{scene-clearing abstraction layer}.
The high-level module (the \textit{Brain}) leverages a Vision–Language Model (VLM) for symbolic task decomposition and distractor identification. It generates structured scene constraints by localizing task-irrelevant elements and producing target-specific segmentation masks. These masks act as domain-invariant geometric priors, isolating spatial regions necessary for manipulation while suppressing background clutter ~\citep{huang2025prism, kossira2025towards}.

The low-level module (the \textit{Cerebellum}) consists of lightweight diffusion-based VLA policies conditioned exclusively on mask-filtered visual observations and proprioceptive feedback. By operating within this task-aligned perceptual subspace, the execution layer focuses on clarified geometric features rather than environmental noise. A central principle of HCS-VLA is strict perception–action consistency: the observation space used during deployment is structurally aligned with that used in training, preventing distractor-induced ambiguity from corrupting policy inference ~\citep{lin2025evo, wen2025llada}. This hierarchical decomposition enables stable sensorimotor execution in dense environments while preserving flexible semantic reasoning for long-horizon tasks.


We validate HCS-VLA on a real bimanual robot in densely cluttered supermarket shelves and in simulation, showing large improvements over strong monolithic VLA baselines in high-density scenes and robust performance on long-horizon tasks such as clutter sorting and restocking.
The contributions of this work are as follows:
\begin{enumerate}
    \item \textbf{Hierarchical Control Architecture:} We propose a structured framework that factorizes end-to-end manipulation into symbolic reasoning and sensorimotor execution, enabling long-horizon orchestration without sacrificing high-frequency responsiveness.
    \item \textbf{Mask-Based Scene Simplification:} We introduce a VLM-guided segmentation mechanism that systematically prunes task-irrelevant distractors, transforming raw observations into geometry-focused representations for robust manipulation.
    \item \textbf{Perception-Action Consistency Protocol:} We establish a principled alignment between offline training and online inference within a clutter-filtered perceptual subspace, demonstrating improved zero-shot robustness and failure recovery in densely cluttered supermarket environments.
\end{enumerate}
\section{Related Work}

\subsection{Vision-Language-Action (VLA) Foundations}

VLA models aim to unify perception and control within a shared transformer-based architecture ~\citep{rt22023arxiv, kim2024openvla}. Early systems such as RT-2 primarily leveraged large-scale web data to improve policy generalization through transfer learning. More recent models, including pi0 ~\citep{black2024pi0} and RDT-1B ~\citep{liu2024rdt}, extend this paradigm to more complex robotic settings, particularly high-degree-of-freedom bimanual manipulation. For example, RDT-1B models multimodal action distributions using a diffusion-based formulation ~\citep{liu2024rdt}, enabling the generation of diverse and coordinated actions.

Despite these advances, existing monolithic VLA models remain sensitive to visual distribution shifts, which limits their robustness when deployed in real-world environments ~\citep{robustness2025, zhou2025exploring}. In this work, we address this limitation by introducing an explicit abstraction layer that separates control reasoning from raw visual inputs, thereby improving robustness to visual variations.

\subsection{Hierarchical Architectures in Robotics}

Hierarchical control has long been a common strategy in robotics, typically decomposing the system into high-level task planning and low-level motion execution modules ~\citep{roboos2025, hwm2026}. Recent frameworks such as RoboOS advocate centralized reasoning to coordinate tasks across multiple embodiments and robotic platforms ~\citep{roboos2025}. Similarly, approaches such as Hi Robot ~\citep{shi2025hi} and VISTA ~\citep{vista2026} employ vision-language models (VLMs) as high-level planners that generate intermediate subgoals for downstream policies.

Our approach differs from these methods in how the interface between the planning and control layers is defined. Instead of relying on pixel-level representations or natural-language subgoals, we introduce a geometric mask as the intermediate representation. This representation provides a more stable and structured signal for downstream control, reducing ambiguity and improving consistency compared to previously explored hierarchical VLA interfaces ~\citep{roboground2025, obeyed2025}.

\subsection{Representation Learning for Manipulation}

Learning appropriate environment representations is crucial for efficient and reliable robotic manipulation. Prior work has explored object-centric representations and geometric primitives to simplify downstream control learning ~\citep{dexgraspvla}. These representations help reduce the complexity of raw visual observations and provide more structured inputs for policy learning.

In practical manipulation scenarios, especially in cluttered environments such as retail shelves or storage spaces, the presence of irrelevant objects and background textures can introduce significant visual distractions for learning-based policies. Instead of relying on raw images, mask-conditioned representations provide a simple yet effective mechanism to isolate task-relevant objects from surrounding clutter.

Building on this idea, our approach explicitly filters the scene using geometric masks that highlight only the objects involved in the manipulation task. By removing irrelevant visual information, the policy can focus on task-critical spatial relationships, which simplifies learning and improves the stability of bimanual manipulation across different object arrangements and shelf layouts.

\section{Problem Formulation}

\subsection{Observations and Geometry-Centric Abstraction}
At each time step $t$, the robot receives an observation $O_t = \{I_t, s_t\}$, where $I_t$ denotes the RGB image and $s_t \in \mathbb{R}^{14}$ denotes the proprioceptive state. The 14-dimensional vector corresponds to the robot’s action space, comprising two 6-DoF arms and one gripper control dimension for each arm. We model the image as a composition of task-relevant geometry and environment-dependent disturbance:
\begin{equation}
I_t = \Phi(\Gamma_t, \xi_t),
\end{equation}
where $\Gamma_t$ represents the geometric structure that is essential for manipulation, and $\xi_t$ summarizes nuisance factors such as clutter, lighting variation, and reflection.

Our goal is not to control the robot directly from raw pixels, but to construct a geometry-preserving visual input that suppresses these disturbances. To this end, we introduce a filtering function $\hat{I}_t = \Psi(I_t)$, where $\hat{I}_t$ retains the spatial structure that is relevant to the task while attenuating task-irrelevant visual content. This abstraction allows the downstream controller to focus on stable geometric cues instead of scene-specific appearance.

\subsection{Hierarchical Policy Structure}
We formulate the manipulation policy as a hierarchical system with two coupled levels. The high-level policy, denoted by $\pi_{\mathrm{H}}$, takes the language instruction $Z$ and the current observation history as input, and produces a sequence of executable subgoals, denoted by $\mathcal{P} = \{g_1, g_2, \dots, g_N\}$, together with spatial constraints that indicate which regions of the scene should be ignored during execution.

The low-level policy, denoted by $\pi_{\mathrm{L}}$, is responsible for continuous control. At time step $t$, it maps the filtered observation $\hat{I}_t$, the proprioceptive state $s_t$, and the active subgoal $g_i$ to a motor action:
\begin{equation}
a_t = \pi_{\mathrm{L}}(\hat{I}_t, s_t, g_i).
\end{equation}
This decomposition separates semantic reasoning from motor generation: the high-level policy decides \emph{what} should be done next, while the low-level policy determines \emph{how} to physically execute it.

\subsection{Action Space and Learning Objective}
The action space $\mathcal{A} \subset \mathbb{R}^{14}$ consists of joint-position increments for two 6-DoF arms and the states of the two grippers. Instead of predicting a single action at each step, the policy outputs an action chunk $\mathbf{A}_t = \{a_t, a_{t+1}, \dots, a_{t+H}\}$ with horizon $H$, which improves temporal smoothness and short-term execution stability.

The low-level controller is trained to maximize task success across environments drawn from a domain set $\mathcal{D}$:
\begin{equation}
\max_{\pi_{\mathrm{L}}} \;
\mathbb{E}_{D \sim \mathcal{D}, \, \tau \sim \pi}
\left[
\sum_{t=0}^{T} \gamma^t \, R_{\mathrm{geo}}(s_t, a_t)
\right],
\end{equation}
where $R_{\mathrm{geo}}$ measures geometric progress toward successful task completion.

To encourage robustness to environmental variation, we impose a domain-invariance requirement on the execution policy:
\begin{equation}
\forall \xi_a, \xi_b,\quad
\pi_{\mathrm{L}}\bigl(\Psi(I(\xi_a)), s_t, g_i\bigr)
=
\pi_{\mathrm{L}}\bigl(\Psi(I(\xi_b)), s_t, g_i\bigr).
\end{equation}
This condition states that, after visual filtering, the low-level policy should depend primarily on task-relevant geometry rather than environment-specific appearance.

\section{Method}

\begin{figure*}[!t]
    \centering
    \includegraphics[width=0.9\textwidth]{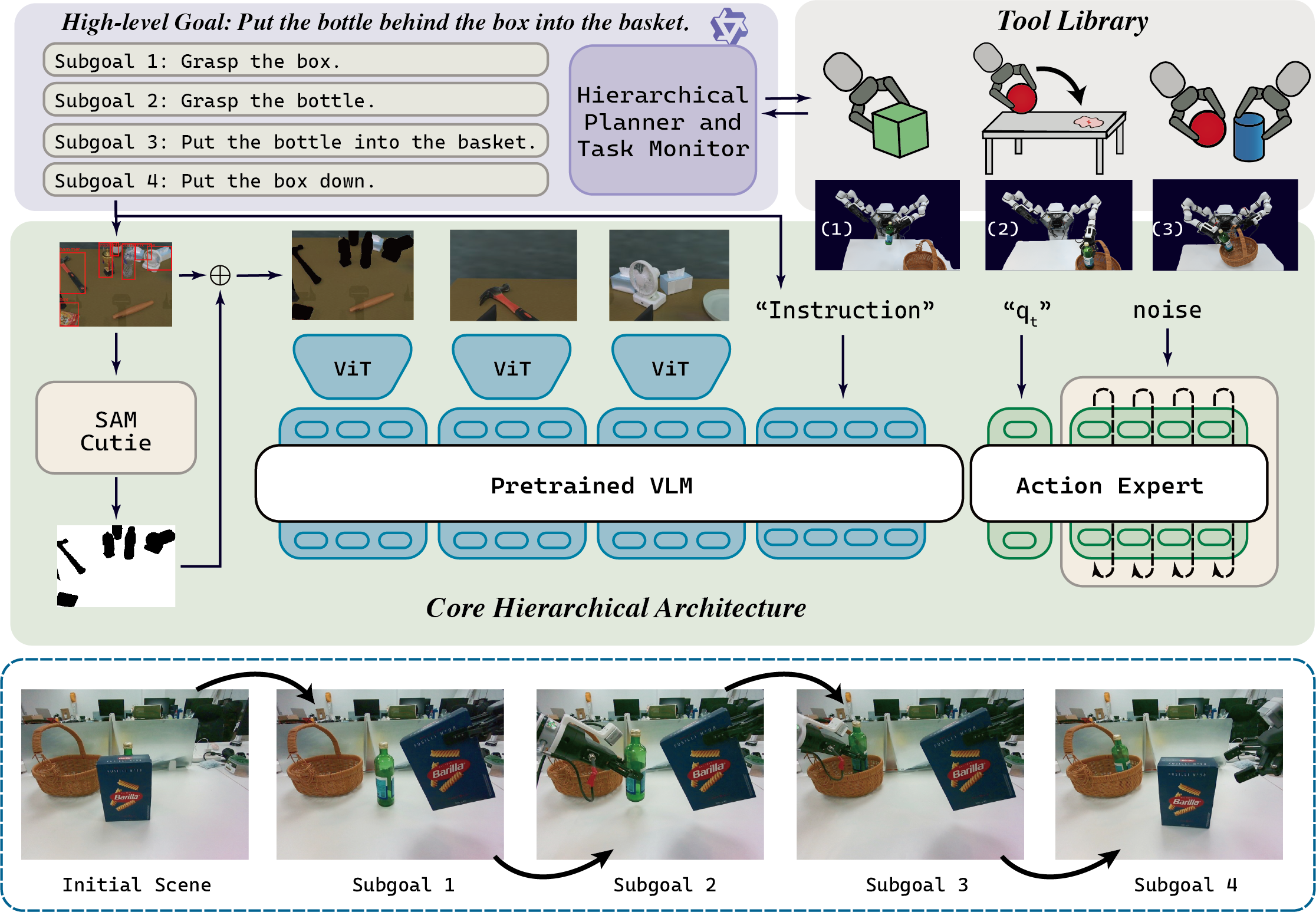}
    \caption{ \textbf{Overview of the HCS-VLA.} Our framework decomposes a high-level natural language goal into a sequence of executable subgoals via a Hierarchical Planner. These subgoals interface with a Tool Library and a vision-based pipeline—integrating SAM/Cutie for segmentation and a Pretrained VLM—to provide multimodal instructions to an Action Expert for precise robot execution.}
    \label{fig:double}
\end{figure*}

We present a hierarchical vision--language--action framework for long-horizon manipulation in unstructured and densely cluttered environments. The key idea is to separate semantic decision-making from continuous motor control, while keeping the two levels tightly connected through visual feedback and execution outcomes. Instead of learning a single end-to-end policy from language and images to actions, our framework decomposes manipulation into high-level task organization and low-level skill execution. This design improves interpretability, modularity, and robustness, especially when the robot operates in scenes with severe clutter and strong visual variation.

\subsection{Overall Framework}

Our framework contains four closely connected modules: a high-level vision--language planner, a geometry-oriented perception module, a diffusion-based skill library, and a verification mechanism for plan update and recovery. The planner determines the current subgoal and the spatial regions that should be ignored. The perception module converts this decision into a filtered visual observation. The skill policy then generates a short action sequence conditioned on the filtered image, the robot state, and the active subgoal. Finally, a verification module checks whether the subgoal has been completed and decides whether the system should continue, retry, or replan.

\smallskip
\noindent\textbf{High-Level Vision--Language Planning.}
The high-level planner is instantiated with Qwen3-v1-235B-A22B-Instruct, a large instruction-tuned vision--language model. To preserve the broad semantic knowledge acquired during large-scale pre-training, the model is used only at inference time and remains frozen throughout both training and deployment. In our framework, the model does not predict motor commands directly. Instead, it serves as a semantic planner that decomposes a long-horizon task into executable subgoals and provides spatial guidance for downstream control.

Given a language instruction $Z$ and the observation history $\mathcal{O}_t = \{I_0, I_1, \dots, I_t\}$, the planner incrementally constructs and updates the subgoal sequence $\mathcal{P}$. Each subgoal $g_i$ is associated with one primitive policy from a tool library $\mathcal{T} = \{\pi_1, \pi_2, \dots, \pi_K\}$, which grounds language-based decisions in a finite set of executable skills.

At each decision step, the planner selects the active subgoal and predicts a set of spatial constraints:
\begin{equation}
(g_t^\star, \mathcal{B}_t)
=
\arg\max_{g, \mathcal{B}}
P(g, \mathcal{B} \mid \mathcal{O}_t, Z, \mathcal{H}_t),
\end{equation}
where $\mathcal{H}_t$ denotes the execution history, including completed subgoals and previous interaction outcomes. The constraint set $\mathcal{B}_t$ contains 2D bounding boxes, where each box is written as $b_i = [x_{\min}, y_{\min}, x_{\max}, y_{\max}]$.

These boxes are not used to mark the target object directly. Instead, they identify task-irrelevant objects or regions that may interfere with the current subtask. In this way, the planner performs top-down scene simplification before low-level control. As a result, the execution policy does not need to interpret the full scene from scratch at every step, and can focus on the geometric structure that matters for manipulation.

\smallskip
\noindent\textbf{Geometry-Oriented Perception by Masked Filtering.}
Real-world manipulation in retail-like scenes often involves heavy clutter, severe occlusion, specular reflection, and changing illumination. If the controller is conditioned directly on raw RGB images, these factors can introduce strong domain shift and reduce execution stability. To address this issue, we apply an explicit masking operation before policy inference.

When a new subgoal starts at time $t_0$, the planner first predicts the bounding boxes $\mathcal{B}_{t_0}$. These boxes are then passed to a zero-shot segmentation model $\mathcal{S}$ to obtain pixel-level masks. We define the task-irrelevant mask as
\begin{equation}
Q_{t_0}(u,v)
=
\mathbb{I}\!\left[
\exists b_i \in \mathcal{B}_{t_0}
\text{ such that }
(u,v) \in \mathcal{S}(I_{t_0}, b_i)
\right],
\end{equation}
where $\mathbb{I}[\cdot]$ is the indicator function.

Running zero-shot segmentation at every frame is computationally expensive and may produce temporally inconsistent masks. Therefore, after initialization, we propagate the mask with a lightweight temporal update module:
\begin{equation}
Q_t = \mathcal{K}(I_t, Q_{t-1}), \qquad t > t_0.
\end{equation}

The filtered image is then computed as
\begin{equation}
\hat{I}_t = \Psi(I_t) = I_t \odot (1 - Q_t),
\end{equation}
where $\odot$ denotes element-wise multiplication. This operation suppresses distractors while preserving the geometry of the regions that are relevant to the current subtask. As a result, the skill policy receives a cleaner and more stable visual input, which improves both cross-domain generalization and control consistency.

\smallskip
\noindent\textbf{Diffusion-Based Skill Execution.}
Each primitive policy in the tool library is implemented as a diffusion-based visuomotor controller. Given the filtered image $\hat{I}_t$, the proprioceptive state $s_t$, and an embedding of the active subgoal $g_t^\star$, the controller generates the action chunk $\mathbf{A}_t$. Compared with one-step autoregressive action prediction, chunk-level generation provides smoother motion and is less sensitive to short-term perceptual noise.

Action generation is modeled as a reverse diffusion process:
\begin{equation}
\mathbf{A}_t^{k-1}
=
\mu_\theta(\mathbf{A}_t^k, k, \hat{I}_t, s_t, g_t^\star)
+
\sigma_k \epsilon,
\end{equation}
where $k$ is the diffusion step, $\mu_\theta$ is a conditional denoising network, and $\sigma_k \epsilon$ is the scheduled noise term. Starting from Gaussian noise, the model gradually refines the trajectory into an action sequence that is consistent with the visual scene, the robot state, and the current semantic goal.

\smallskip
\noindent\textbf{Verification and Adaptive Replanning.}
Because physical interaction is inherently uncertain, the system must continually evaluate execution outcomes. After each action chunk, or at predefined checkpoints, we apply a binary verification function $\mathcal{V}(I_t, g_t^\star) \in \{0,1\}$, which determines whether the current subgoal has been completed successfully.

If the subgoal is achieved, the planner removes it from the remaining plan:
\begin{equation}
\text{if } \mathcal{V}(I_{t+H}, g_t^\star)=1,
\qquad
\mathcal{P} \leftarrow \mathcal{P} \setminus \{g_t^\star\}.
\end{equation}

If verification fails, the system does not terminate immediately. Instead, the planner updates the execution strategy based on the observed failure. This update may keep the current skill and retry the same subgoal, revise the spatial constraints to reveal newly important regions, or modify the remaining subgoal sequence when the original plan is no longer suitable. In this way, high-level reasoning and low-level control remain coupled throughout the task, enabling robust execution in dynamic and cluttered environments.

\subsection{Data Collection}
To train low-level policies that remain effective in realistic retail settings, we collected a multi-task manipulation dataset directly in indoor supermarket shelf environments. The scenes are densely cluttered with everyday retail products, including common grocery and household items with diverse shapes, sizes, textures, and packaging appearances. This setup exposes the robot to realistic visual variation, partial occlusion, and constrained manipulation space, which are common in practical retail operations.

The dataset contains 2,100 complete expert trajectories collected through kinesthetic teaching on a real dual-arm robotic platform. Each trajectory is written as $\tau = \{(I_t, s_t, a_t)\}_{t=0}^{T}$, where the RGB image, proprioceptive state, and control action are recorded synchronously. The demonstrations cover three representative manipulation categories: single-arm grasp-and-stabilize with either the left or right arm, single-arm placement of an item to a specified target location, and bimanual cooperative grasping for objects that require coordinated two-hand interaction. Together, these tasks provide supervision for both unilateral and bilateral manipulation behaviors in shelf-based retail scenarios.

A key principle in our data pipeline is to keep the visual preprocessing used during training consistent with that used at test time. To achieve this, we build an automatic offline annotation pipeline that mirrors the online planning process. For each demonstration, the raw RGB sequence is processed by the same planner stack used during deployment. Specifically, the vision--language planner first predicts task-irrelevant bounding boxes in the initial frame. These boxes are then refined by the segmentation model and propagated over time by the mask update module, resulting in a frame-aligned mask sequence $\{Q_t\}_{t=0}^{T}$.

The final training set is constructed from samples of the form
\begin{equation}
x_t = (\hat{I}_t, s_t, g, \mathbf{A}_t),
\end{equation}
where $\hat{I}_t = I_t \odot (1 - Q_t)$ and $\mathbf{A}_t = \{a_t, \dots, a_{t+H}\}$. The resulting dataset is
\begin{equation}
\mathcal{D} = \{x_t\}_{t=0}^{T}.
\end{equation}
Using the same planner, segmentation model, and temporal mask propagation strategy in both offline data construction and online execution reduces the distribution gap caused by inconsistent visual preprocessing.

\section{Experiments}

We evaluate HSC-VLA in densely cluttered supermarket environments, with an emphasis on whether explicit scene clearing improves reliability when targets are surrounded or partially occluded by distractors. Our experiments focus on three aspects. First, we test whether hierarchical scene simplification improves success rates in high-clutter settings compared to strong monolithic baselines. Second, we analyze how performance changes across grasping, placing, and bimanual manipulation as clutter increases. Third, we examine whether keeping a simplified scene representation throughout execution improves stability in long-horizon tasks.

\subsection{Experimental Setup}

\leavevmode
\begin{wrapfigure}{r}{0.45\columnwidth}
    \centering
    \includegraphics[width=0.45\columnwidth]{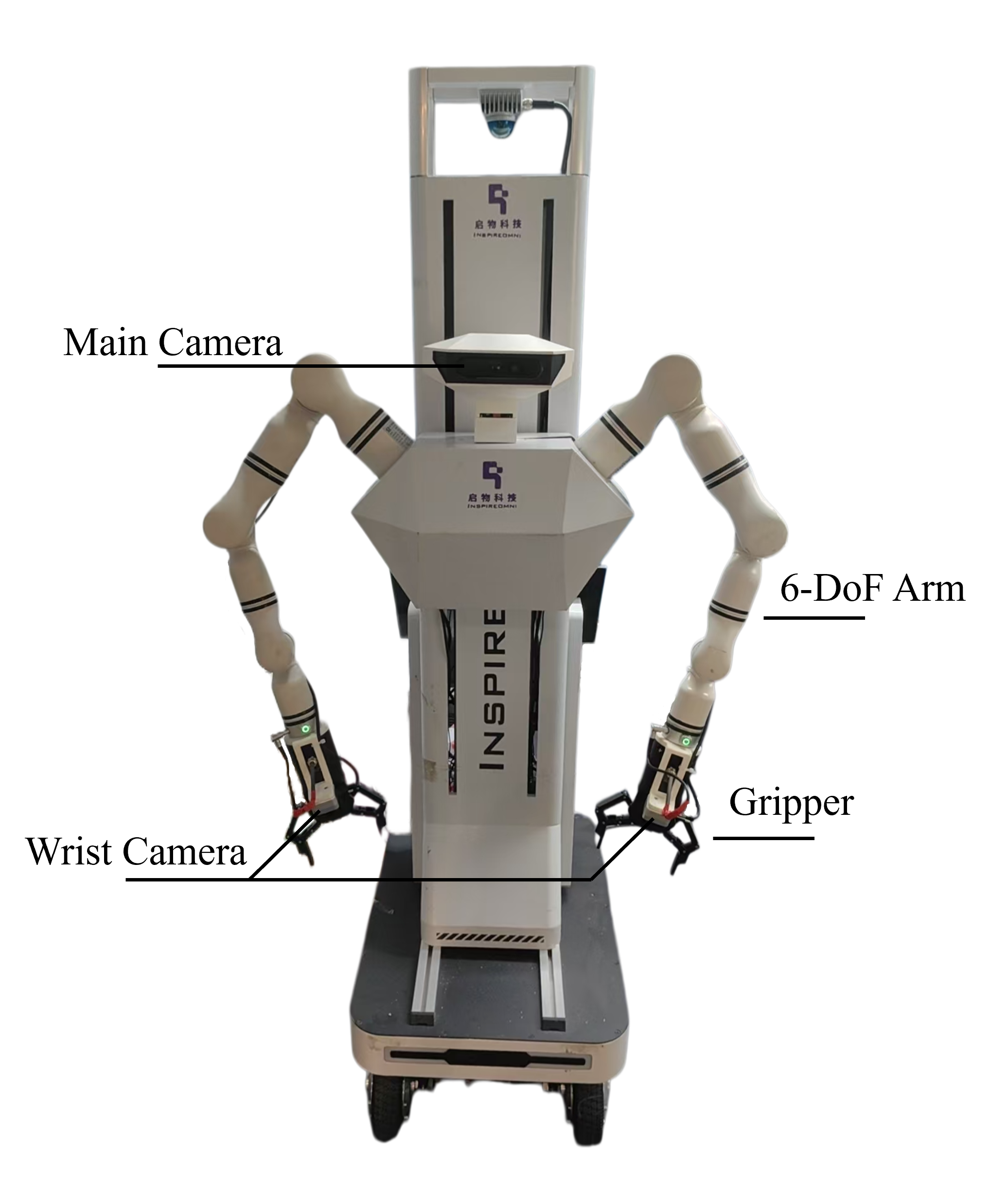}
    \vspace{-4mm}
    \caption{Robot system used in real-world evaluations.}
    \vspace{-4mm}
    \label{fig:robot}
\end{wrapfigure}

\noindent\textbf{Hardware and environment:} We deploy on the InspireOmni O1 bimanual platform with two 6-DoF arms and parallel grippers. The workspace mimics supermarket shelves with extreme clutter, typically 5 to 20 objects per shelf section. This setting produces frequent occlusions and strong visual distraction.

\smallskip\noindent\textbf{Simulation environment:} We additionally benchmark atomic skills in RoboTwin 2.0, a physics-based suite for robust bimanual manipulation with standardized evaluation and domain randomization~\citep{chen2025benchmarking}. To align with the three skill categories in Table~\ref{tab:main_results}, we select \textit{Adjust Bottle}, \textit{Grab Roller}, and \textit{Lift Pot} as grasping tasks; \textit{Place Burger Fries} and \textit{Click Alarmclock} as placing tasks; and \textit{Handover Mic} as a bimanual interaction task. Together, these tasks cover diverse occlusion patterns, contact dynamics, and precision requirements while matching our grasp, place, and bimanual abstraction.

\smallskip\noindent
\textbf{Training details:} Across methods, we use batch size 32 and train for 50k gradient steps without early stopping. We adopt AdamW with a step-based schedule. We warm up linearly for 1k steps to a peak learning rate of $5\times10^{-5}$, then apply cosine decay to $2.5\times10^{-6}$ over the remaining steps. Unless stated otherwise, we report performance from the final checkpoint.

\smallskip\noindent
\textbf{Task suite:}
\begin{itemize}
    \item \textit{Toolbox tasks}: single-step primitives that cover precision grasping, target placing, and bimanual coordinated handling. In simulation, these correspond to the RoboTwin 2.0 tasks listed above. In the real supermarket setup, we evaluate the same three primitive types under low-density and high-density clutter as shown in Table~\ref{tab:main_results}.
    \item \textit{Clutter sorting}: a multi-step setting where the robot rearranges distractors to expose and extract a target object. Success requires explicit clutter management in addition to local control.
    \item \textit{Restocking}: a long-horizon sequence where the robot identifies an empty slot, picks an item from a bin, and places it onto the shelf. This task stresses robustness under sequential decision making.
\end{itemize}

\subsection{Comparison with Baselines}

\begin{table*}[t]
    \centering
    \caption{Experimental evaluation of HSC-VLA. We compare our method with representative VLA baselines on grasping tasks in scenes with different clutter densities, analyze the effect of scene simplification strategies, and evaluate robustness on long-horizon manipulation tasks. \textbf{Bold} indicates the best performance, and \textcolor{cyan}{cyan} indicates the second-best performance. SR@$n$ denotes the success rate over $n$ trials.}
    \label{tab:comprehensive_results}
    
    \vspace{0ex}

    \begin{subtable}{\textwidth}
        \centering
        \caption{Main results: success rate (\%) under different clutter densities.}
        \label{tab:main_results}
        \small
        \begin{tabular*}{\textwidth}{@{\extracolsep{\fill}}l cccc cccc @{}}
            \toprule
            \multirow{3}{*}{Method} & \multicolumn{4}{c}{Low Density} & \multicolumn{4}{c}{High Density} \\
            \cmidrule(lr){2-5} \cmidrule(lr){6-9}
            & Grasp & Place & Bimanual & \textbf{Aggr.} & Grasp & Place & Bimanual & \textbf{Aggr.} \\
            & {\scriptsize SR@300} & {\scriptsize SR@200} & {\scriptsize SR@100} & & {\scriptsize SR@300} & {\scriptsize SR@200} & {\scriptsize SR@100} & \\ 
            \midrule
            ACT~\citep{zhao2023learning} & 76\% & 41\% & 85\% & 67.3\% & 16\% & 2\% & 0\% & 6.0\% \\
            DP~\citep{chi2025diffusion} & 78\% & 67\% & 53\% & 66.0\% & 0\% & 3\% & 0\% & 1.0\% \\
            DP3~\citep{ze20243d} & \textbf{98\%} & 72\% & 91\% & 87.0\% & 1\% & 16\% & 3\% & 6.7\% \\
            RDT~\citep{liu2024rdt} & 76\% & 56\% & 90\% & 74.0\% & 43\% & \textcolor{cyan}{20\%} & \textcolor{cyan}{31\%} & 31.3\% \\
            $\pi_0$-LoRA~\citep{black2024pi0} & 80\% & \textcolor{cyan}{78\%} & 92\% & 83.3\% & 67\% & 8\% & 13\% & 29.3\% \\
            $\pi_0$-Full FT~\citep{black2024pi0} & 90\% & \textcolor{cyan}{78\%} & \textcolor{cyan}{95\%} & \textcolor{cyan}{87.7\%} & \textcolor{cyan}{75\%} & 13\% & 15\% & \textcolor{cyan}{34.3\%} \\
            \textbf{Ours: HSC-VLA} & \textcolor{cyan}{92\%} & \textbf{84\%} & \textbf{96\%} & \textbf{90.7\%} & \textbf{85\%} & \textbf{78\%} & \textbf{97\%} & \textbf{86.7\%} \\ 
            \bottomrule
        \end{tabular*}
        \vspace{1ex}
    \end{subtable}

    \makebox[\textwidth][c]{%
        \hspace{0.05\textwidth}
        \begin{subtable}{0.48\textwidth}
            \centering
            \caption{Ablation on scene simplification.}
            \label{tab:ablation}
            \small
            \begin{tabular}{@{}lccc@{}}
                \toprule
                \multirow{2}{*}{Strategy} & Low Density & High Density & Long-Hori. \\
                & {\scriptsize SR@100} & {\scriptsize SR@100} & {\scriptsize SR@50} \\ 
                \midrule
                Base VLA, no mask & \textcolor{cyan}{90\%} & 56\% & \textcolor{cyan}{40\%} \\
                Static mask & \textbf{98\%} & \textcolor{cyan}{69\%} & 10\% \\
                \textbf{Dynamic clearing} & \textbf{98\%} & \textbf{80\%} & \textbf{72\%} \\ 
                \bottomrule
            \end{tabular}
        \end{subtable}
        \hspace{0.05\textwidth}
        \begin{subtable}{0.48\textwidth}
            \centering
            \caption{Robustness on long-horizon tasks.}
            \label{tab:robustness}
            \small
            \begin{tabular}{@{}lccc@{}}
                \toprule
                Task & Steps & Best Base. & \textbf{Ours} \\ 
                & & {\scriptsize SR@50} & {\scriptsize SR@50} \\ 
                \midrule
                Toolbox tasks & 1 & 88\% & 90\% \\
                Clutter sorting & $\geq 3$ & 40\% & 72\% \\
                Restocking & 5 & 14\% & 66\% \\ 
                \bottomrule
            \end{tabular}
        \end{subtable}
    }
\end{table*}

We compare HSC-VLA with representative baselines spanning behavior cloning, diffusion-based policies, and vision--language action models, including ACT~\citep{zhao2023learning}, DP~\citep{chi2025diffusion}, DP3~\citep{ze20243d}, RDT~\citep{liu2024rdt}, and $\pi_0$ under LoRA or full fine-tuning~\citep{black2024pi0}. Table~\ref{tab:main_results} summarizes results.

\textbf{Low density:} When targets are visually accessible, several baselines remain competitive. DP3 and $\pi_0$-Full FT achieve 87.0\% and 87.7\% aggregate success, while HSC-VLA attains the best overall result at 90.7\%. The largest gains appear in placing and bimanual manipulation, suggesting that explicit abstraction and clearing can still help even when clutter is mild.

\textbf{High density:} Under heavy clutter, monolithic policies degrade sharply, indicating difficulty in maintaining stable grounding amid occlusions and distractors. DP3 drops from 87.0\% aggregate success to 6.7\%, and ACT decreases from 67.3\% to 6.0\%. The strongest baseline in this regime, $\pi_0$-Full FT, reaches 34.3\% aggregate success. In contrast, HSC-VLA maintains 86.7\% aggregate success, with strong performance on grasping, placing, and bimanual manipulation. This corresponds to a 52.4 percentage-point absolute improvement over the strongest baseline.

Task-level results further show that placing and bimanual execution are most sensitive to clutter for monolithic approaches. In high density, the best baseline reaches 20\% on placing and 31\% on bimanual, whereas HSC-VLA reaches 78\% and 97\%. Overall, filtering task-irrelevant clutter before low-level action generation provides a substantially cleaner control interface when success depends on preserving precise, step-specific geometry throughout execution.

\subsection{Ablation: Scene Simplification Strategies}

\begin{figure}[!t]
    \centering
    \begin{subfigure}[c]{0.58\columnwidth}
        \centering
        \includegraphics[width=\linewidth]{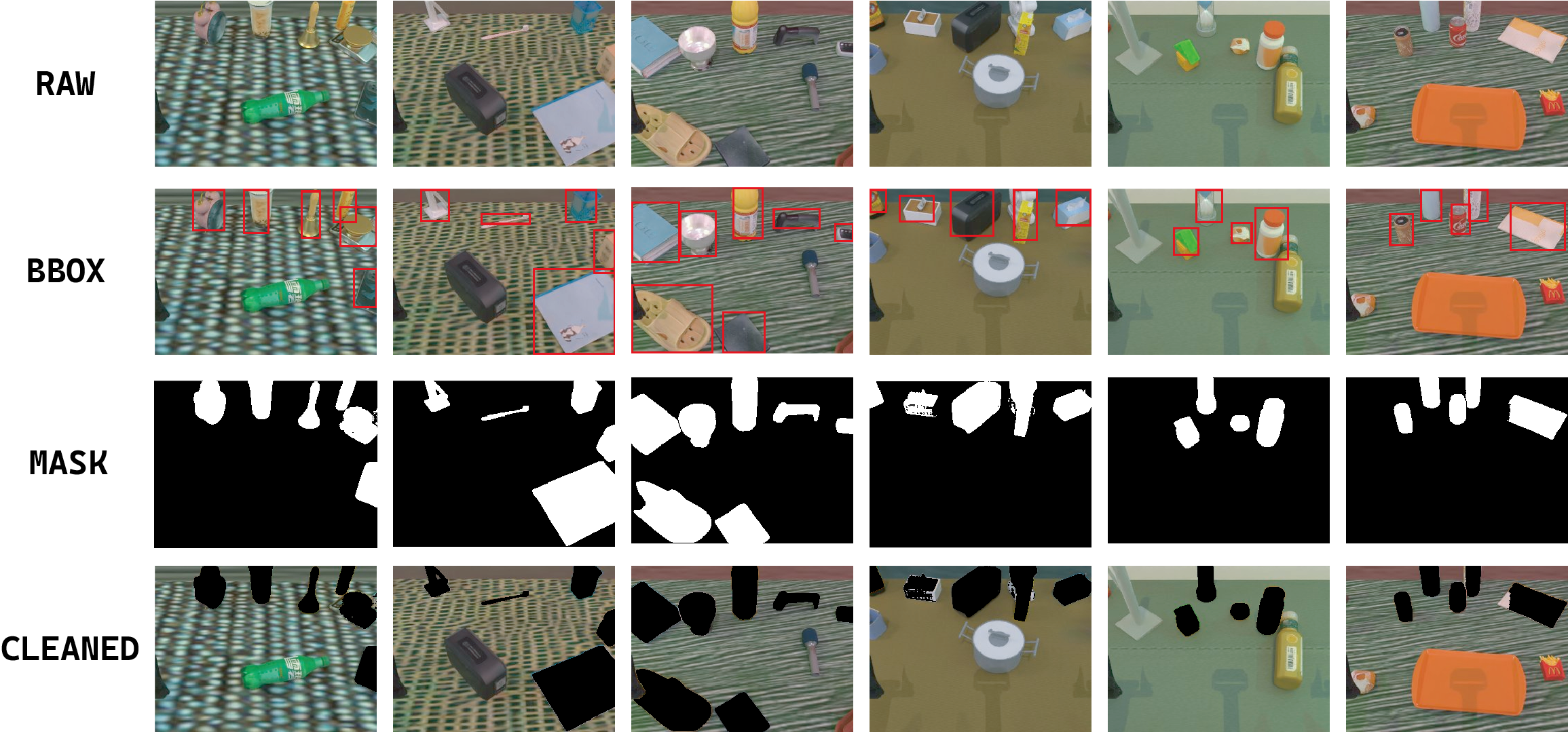}
        \caption{Masking visualization during execution. The policy suppresses task-irrelevant distractors and focuses on target-relevant regions across grasping, placing, and bimanual primitives.}
        \label{fig:processed}
    \end{subfigure}
    \hfill
    \begin{subfigure}[c]{0.38\columnwidth}
        \centering
        \includegraphics[width=\linewidth]{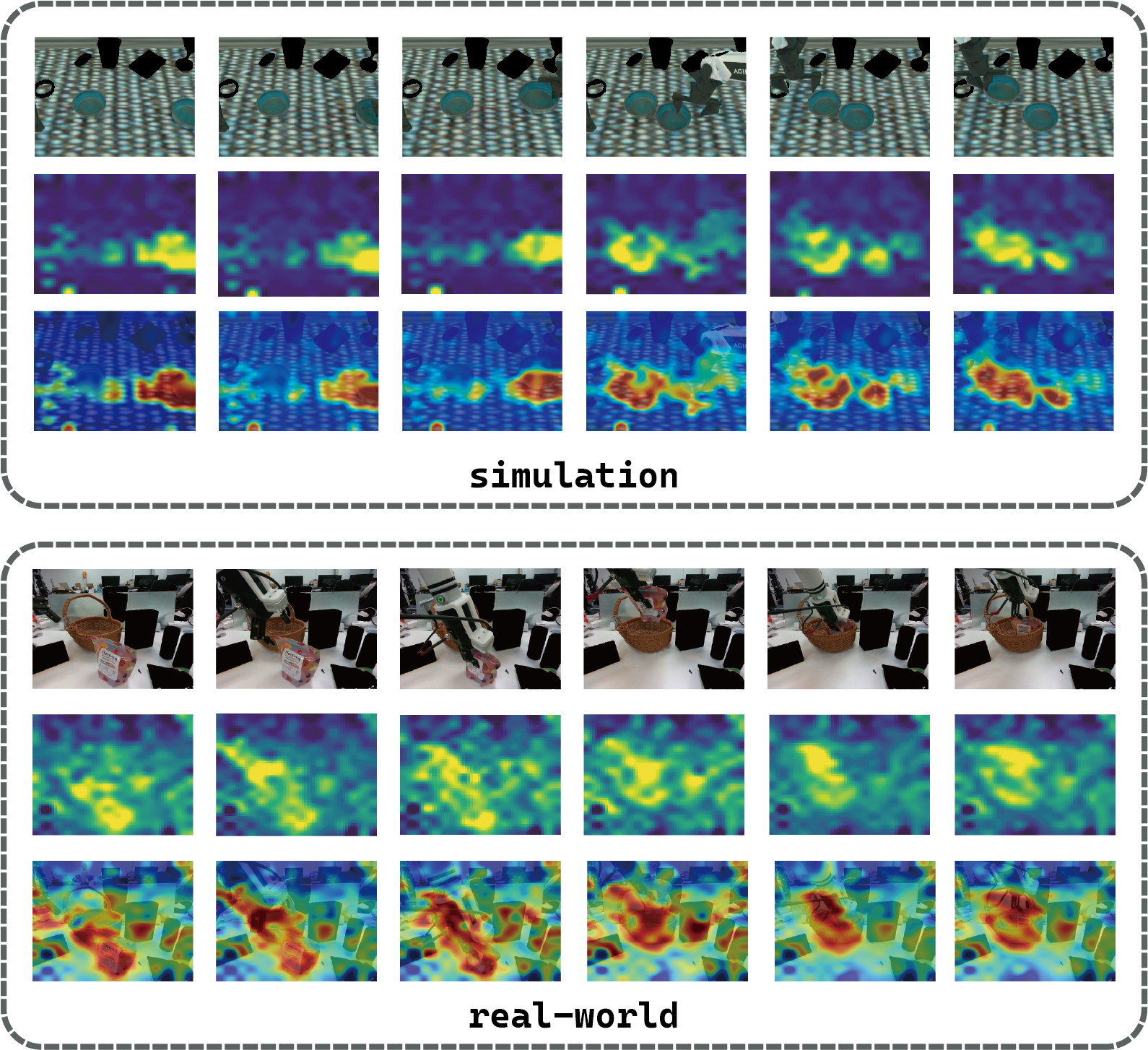}
        \caption{Attention map tracking in simulation and real-world trials. Masking reduces attention drift toward task-irrelevant clutter.}
        \label{fig:attentionmap}
    \end{subfigure}
    \caption{Visualization of scene filtering during task execution. Left: masked observation used by the policy. Right: attention evolution in simulation and real-world trials. Together, these results show that the proposed masking mechanism effectively suppresses irrelevant clutter and improves task-focused perception.}
    \label{fig:mask_attention}
\end{figure}

Figure~\ref{fig:attentionmap} illustrates attention behavior during execution. Without masking, attention can drift to salient but irrelevant clutter, weakening task grounding. Our mask suppresses distractors while preserving task-relevant regions, improving stability under occlusions and frequent scene changes.

We quantify this effect using three variants in Table~\ref{tab:ablation}. The no-mask baseline uses raw observations. The static mask computes a mask once and reuses it throughout execution. Our dynamic clearing updates the mask online as the scene evolves.

For low-density and high-density ablations, we evaluate the same primitive task, uprighting a toppled bottle, while varying surrounding clutter. Under high density, dynamic clearing achieves the best result at 80\% SR@100, outperforming the static mask at 69\% and the no-mask baseline at 56\%. Under low density, both masking variants reach 98\% SR@100, indicating that masking does not hinder execution when occlusion is mild.

We further test long-horizon behavior using a multi-step collection task where the robot collects multiple bowls and places them together. The static mask performs poorly at 10\% SR@50 because the fixed mask becomes outdated as objects move, and it may suppress newly relevant regions. The no-mask baseline improves to 40\% SR@50 but remains sensitive to attention drift. By updating the mask after each interaction, dynamic clearing maintains task-relevant focus across steps and reaches 72\% SR@50.

\subsection{Robustness in Long-Horizon Tasks}
In multi-step scenarios, perception and grounding errors accumulate over time. This effect is amplified in dense clutter where occlusions are frequent and distractors can move after each interaction. As a result, monolithic policies often fail even when individual primitives appear reliable.

Table~\ref{tab:robustness} reports results across three task families. On \textit{Toolbox tasks}, HSC-VLA matches short-horizon performance and slightly improves over the best baseline from 88\% to 90\%. The gain becomes much larger as the horizon increases. For \textit{Clutter sorting}, the best baseline reaches 40\% SR@50, while HSC-VLA achieves 72\%. This indicates that maintaining a simplified scene representation across successive interactions reduces cascading failures caused by attention drift and unstable target localization. For \textit{Restocking}, HSC-VLA improves success from 14\% to 66\%, suggesting that continual clutter suppression and step-wise grounding stabilization can substantially mitigate error accumulation in sequential decision making.

\section{Conclusion}
We presented HSC-VLA, a hierarchical vision--language--action framework for long-horizon manipulation in densely cluttered retail environments. By making scene clearing explicit, a high-level Brain decomposes tasks and generates target-specific masks, while a diffusion-based Cerebellum executes bimanual skills on mask-filtered observations with consistent preprocessing in training and deployment. Experiments on real supermarket shelves show substantial gains over monolithic VLA baselines under high-density clutter, and ablations confirm that dynamic clearing is critical for long-horizon stability.

Despite these improvements, HSC-VLA still has limitations. The pipeline relies on the quality of high-level distractor localization and segmentation, and failures in mask initialization or mask drift can propagate to execution. Dynamic mask updates also introduce extra computation and latency, which can become a bottleneck for fast interactions. In addition, the current tool library constrains the space of behaviors and may limit generalization to tasks that require new primitives or more complex contact dynamics. Future work will improve temporal mask tracking, reduce online overhead, and expand the skill set while preserving perception--action consistency.

\clearpage

\bibliography{main}
\bibliographystyle{bibstyle}



\end{document}